Censoring-Aware Tree-Based Reinforcement Learning for Estimating Dynamic Treatment Regimes with Censored Outcomes

Animesh Kumar Paul[1,2], Russell Greiner[1,2,3]

[1]Department of Computing Science, University of Alberta, Edmonton, AB, Canada
[2]Alberta Machine Intelligence Institute, Edmonton, Canada
[3]Department of Psychiatry, University of Alberta, Edmonton, AB, Canada.

## Abstract

Dynamic Treatment Regimes (DTRs) provide a systematic approach for making sequential treatment decisions that adapt to individual patient characteristics, particularly in clinical contexts where survival outcomes are of interest. Censoring-Aware Tree-Based Reinforcement Learning (CA-TRL) is a novel framework to address the complexities associated with censored data when estimating optimal DTRs. We explore ways to learn effective DTRs, from observational data. By enhancing traditional tree-based reinforcement learning methods with augmented inverse probability weighting (AIPW) and censoring-aware modifications, CA-TRL delivers robust and interpretable treatment strategies. We demonstrate its effectiveness through extensive simulations and real-world applications using the SANAD epilepsy dataset, where it outperformed the recently proposed ASCL method in key metrics such as restricted mean survival time (RMST) and decision-making accuracy. This work represents a step forward in advancing personalized and data-driven treatment strategies across diverse healthcare settings.

# Introduction

Dynamic Treatment Regimes (DTRs) offer a systematic approach to optimizing sequential treatment decisions by tailoring interventions based on individual patient characteristics and evolving clinical histories [1,2]. Widely applicable in managing chronic and progressive diseases, such as cancer and diabetes, DTRs consist of stage-specific decision rules that guide treatment allocation to maximize long-term patient outcomes [3,4]. We consider ways to learn DTRs from observational data. Grounded in causal inference [5] (described in the Notation and Assumptions section under the Methods section below), the primary goal of learning is to produce a DTR model that can personalize treatment strategies by accounting for the heterogeneity in treatment responses, thereby enhancing both individual and population-level health outcomes.

Here, we measure the effectiveness (reward) of a policy based on the survival time of the individual. For datasets with only uncensored information – that is, where survival outcomes are fully observed for all individuals – various statistical and machine learning techniques have been proposed to estimate optimal DTRs. Traditional methods – such as Q-learning [6–8] and A-learning [9,10] – rely on regression-based models to iteratively estimate stage-specific decision rules. While effective, these methods often require strong parametric assumptions and can struggle with scalability in high-dimensional settings. To overcome these limitations, tree-based methods have emerged as a robust alternative due to their flexibility and interpretability. Among these, the Tree-Based Reinforcement Learning (T-RL) algorithm, introduced by Tao et al.[11], embeds augmented inverse probability weighting (AIPW)[12] into the tree structure to estimate optimal treatment rules across multiple stages. T-RL has demonstrated excellent performance on complete datasets, particularly due to its ability to handle complex interactions between covariates and treatments while producing interpretable decision rules, leveraging the inherent transparency and simplicity of decision tree structures.

In the context of censored data, methods often integrate survival analysis techniques to address incomplete observations – in particular, situations where we want to know when a training individual will die, but know only that this patient is still alive 2 years later after a specific baseline time point (eg, the start of the study), meaning that patient lived at least 2 years. Approaches such as inverse probability censoring weighting (IPCW)[13–16] and pseudo-value-based methods[17] have been developed to adjust for such censoring in survival outcomes. Pseudo-value methods, for example, replace censored survival quantities with derived pseudo-observations to allow us to apply standard machine learning techniques to such datasets with censored individuals. These pseudo-observations approximate the target quantity, such as restricted mean survival time, for both censored and uncensored individuals, using information derived from survival models like the Cox proportional hazard model. Despite advancements in handling censored data, no tree-based methods have yet been extended to incorporate censoring directly into the estimation framework for DTRs.

To address this gap, we propose **Censoring-Aware Tree-Based Reinforcement Learning (CA-TRL)**, an extension of the T-RL method tailored for right-censored data, where the exact timing of an event is unknown but is known to occur after a certain observed time point. Building on the robustness and interpretability of T-RL, CA-TRL introduces censoring adjustments through a modified AIPW estimator that incorporates survival and censoring probabilities into the tree-building process. This ensures unbiased estimation of counterfactual outcomes while retaining the flexibility of tree-based methods. By leveraging T-RL's strengths and addressing its limitations in handling incomplete data, CA-TRL provides a scalable and interpretable solution for optimizing multi-stage, multi-treatment decision-making under censoring conditions.

The remainder of this paper is organized as follows. The Methods section details the notation and assumptions underlying the CA-TRL framework, and its methodology, including the modified purity measures and recursive estimation process. The Results section evaluates the performance of CA-TRL through simulation studies and a real-world dataset, comparing it to baseline models including fixed treatment policies (uniform treatment assignment), random treatment assignment policies, and a recent advanced DTR method, demonstrating its utility in clinical decision-making under censored data conditions. Finally, the Discussion section concludes with a discussion of the method's implications and potential extensions for future research.

# Results

To thoroughly assess the performance of the proposed method, we performed extensive experiments, of our method, baseline models including fixed treatment policies (uniform treatment assignment), random treatment assignment policies, and a recent advanced DTR method, ASCL[17], on a semi-synthetic and real-world datasets, using four distinct evaluation metrics, as described in the Methods subsection. Given the absence of a true reward function or true policy in existing survival datasets, we generated semi-synthetic datasets by integrating real-world data with simulated censoring mechanisms. Importantly, the survival models only utilize the observed censored times for the subjects affected by those simulated censoring mechanisms. The true underlying information, including the given reward function and true policy used for generating the semi-synthetic data, was exclusively used during the evaluation phase to assess the learned policy's effectiveness and was not incorporated during the training process.

### i. Semi-Synthetic Data Study

Given the challenges of directly validating learned policies due to the absence of true reward functions and corresponding optimal policies in real-world survival datasets, we created a semi-synthetic dataset. This dataset preserved the structure and characteristics of real-world data, including realistic covariate distributions and censoring percentages, while incorporating simulated outcomes and treatment assignments to ensure a controlled yet practical evaluation environment.

For this study, we utilized the MIMIC-IV v2.0 dataset as the primary source [18]. Coronary artery disease (CAD) was selected as the focus due to its prevalence and significance as a leading cause of hospitalization and mortality worldwide. The dataset provided comprehensive clinical information, including demographics, laboratory measurements, and outcomes from hospital admissions. To refine the analysis, we focused on non-ICU admissions, treating each hospital admission as an independent patient record to capture diverse clinical trajectories.

A 48-hour observation window was defined, divided into two 24-hour intervals. For the experimental purpose, we have considered the first interval as the baseline covariates for Stage 1 (renal and anemia management), while the second interval as time-dependent changes for Stage 2 (glycemic management). The features included in the analysis were Age, Creatinine, Hemoglobin, Potassium, Sodium, Glucose, Platelet Count, Hematocrit, and White Blood Cell Count. Survival time was measured from admission to death or discharge, with approximately 62.18% of CAD patients exhibiting censored survival times. This censoring rate was replicated in the semi-synthetic dataset to maintain realistic evaluation conditions. For the initial preprocessing, we employed the recently published data processing pipeline for MIMIC-IV by Mehak et al. [19]. After preprocessing, a total of 65,486 patient records were available for analysis.

Data Generation

The processed data were used to simulate a two-stage DTR problem. Stage 1 focused on renal and anemia management using baseline covariates, while Stage 2 addressed glycemic control. The primary goal was to identify treatment policies that maximize survival outcomes. While we aimed to align the reward function, true policy, and other simulation-defined functions with clinical reasoning, we acknowledge that these may not fully represent the complexities of individual CAD treatment. Hence, the following functions should not be considered as ground truth for clinical decision-making.

At Stage 1, the treatment indicator $A_1$ was assigned using a multinomial distribution. For binary ($M = 2$) and three-categorical ($M = 3$) treatment options, the treatment probability vector was defined as follows: for binary treatments, the reference treatment probability ($\pi_{10}$) was set to 1, and the alternative treatment probability $\pi_{11}$ was modeled as $exp(0.2 * Creatinine_1 + 0.2 * Hemoglobin_1)$. For three-categorical treatments, the reference treatment probability remained 1, while the additional treatment probabilities ($\pi_{11}$ and $\pi_{12}$) were modeled as $exp(0.2 * Creatinine_1 + 0.1 * Potassium_1)$ and $exp(0.2 * Hemoglobin_1 - 0.02 * Age)$, respectively. In both binary and three-categorical cases, these probabilities were then normalized to ensure they sum to 1 and used to simulate treatment assignments for each subject. Importantly, these represent treatment propensity models and do not correspond to censoring probabilities, which are separately modeled and incorporated into the data generation process. The optimal treatment rule at Stage 1, $g_1^{opt}$, was determined using clinically relevant covariates. For binary treatments, the optimal treatment rule was

$$g_1^{opt} = I(Creatinine_1 > 1.5) * I(Hemoglobin_1 \leq 12) \quad \in \{0, 1\},$$

while for three-categorical treatments, it was

$$g_1^{opt} = I(Creatinine_1 > 1.5) * (1 + I(Hemoglobin_1 \leq 10)) \quad \in \{0,1,2\}.$$

For our simulation, we generated the Stage 1 outcome $T_1$ (which was the time until the next decision point, which could be either another treatment stage or terminating the process—perhaps due to an event like kidney failure, or death) using the equation $exp(1.5 + 0.3 * Potassium_1 - |1.5 * Creatinine_1 - 2| * (A_1 - g_1^{opt})^2 + \varepsilon)$, where ε ~ Exp( λ ) represents a noise term added to capture variability in survival times.

In Stage 2, treatment assignments ($A_2$) were generated based on updated covariates, including the first-stage outcome ($T_1$). $A_2$ was assigned using a multinomial distribution. For binary treatments, the reference treatment (unnormalized) probability ($\pi_{20}$) remained fixed at 1, while the alternative treatment unnormalized probability was $\pi_{21} = exp(0.002 * Glucose_2 + 0.005 * T_1)$. For three-categorical treatments, the unnormalized probabilities were $\pi_{21} = exp(0.002 * Glucose_2 + 0.002 * Sodium_2)$ and $\pi_{22} = exp(0.005 * Platelet\ Count_2 + 0.005 * T_1)$. The optimal treatment rule at the second stage ($g_2^{opt}$) was defined similarly. For binary treatments, the optimal treatment rule was

$$g_2^{opt} = I(Glucose_2 > 140) + I(T_1 < 3),$$

while for three-categorical treatments, it was

$$g_2^{opt} = I(Glucose_2 > 140) * (I(T_1 > 0.5) + I(T_1 > 3)).$$

The second-stage outcome ($T_2$) was generated using the function

$$T_2 = exp(1.18 + 0.2 * T_1 - |0.5 * Glucose_2 + 2| * (A_2 - g_2^{opt})^2 + \varepsilon).$$

The methodology for defining treatment probabilities, optimal treatment assignment, and outcome functions follows recent works on synthetic data generation for evaluating DTR models [11,17], ensuring consistency with state-of-the-art approaches. For the optimal treatment rule and outcome (or reward) function, we selected key features based on their well-established clinical relevance. For Stage 1, Creatinine (>1.5 mg/dL) was chosen because elevated levels indicate significant renal impairment and are strongly linked to increased cardiovascular risk[20], while low Hemoglobin is a marker for anemia, which adversely affects cardiac outcomes[21–23]. For Stage 2, we use a Glucose threshold of >140 mg/dL to identify hyperglycemia that necessitates more aggressive glycemic management[24]. In addition, routinely measured features such as Age, Potassium, Sodium, Platelet Count, Hematocrit, and White Blood Cell Count are included to capture a comprehensive cardiovascular risk profile[25–30].

In our survival outcome function, the exponential transformation is employed to ensure that the predicted survival times remain strictly positive—a fundamental requirement in time-to-event analyses. This transformation is standard practice in survival analysis, as seen in both the Cox proportional hazards model and accelerated failure time models, where exponential forms are used to capture the multiplicative effect of covariates on the baseline hazard[31,32]. Moreover, the quadratic penalty term $(A - g^{opt})^2$ is incorporated to measure the deviation of the actual treatment $A$ from the optimal treatment $g^{opt}$. Squaring this difference ensures that the penalty is symmetric and increases nonlinearly with larger deviations, reflecting the intuition that departures from the optimal decision should incur disproportionately higher costs. This approach is well supported by literature in DTRs for generating the synthetic data[11,17].

Censoring mechanisms were incorporated into the data generation process to mimic real-world survival datasets, where complete observation of survival times is often unavailable. The censoring times ($C$) were generated using three distinct distributions to reflect varying levels of censoring:

1. **Exponential Censoring**: Censoring times were drawn from an exponential distribution, $C \sim Exp(c_0)$, where $c_0$ was chosen appropriately to achieve the desired censoring rate.
2. **Conditional Exponential Censoring**: To introduce covariate dependency, censoring times followed $C \sim c_0 * exp(0.3 \cdot Creatinine_1 + 0.2 \cdot |Potassium_1 - 4.0|)$.
3. **Uniform Censoring**: Censoring times were generated from a uniform distribution, $C \sim Unif(a, b)$, where $a$ and $b$ define the lower and upper bounds, respectively, and were chosen appropriately to achieve the desired censoring rate.

These distributions were calibrated to achieve a censoring rate of around 62%, consistent with the MIMIC CAD population.

To determine whether a subject proceeds to the second stage, the binary indicator variable $\eta = I(T_1 < C)$ was used, where $\eta = 1$ indicates that the first-stage outcome ($T_1$) is less than the censoring time (C), allowing progression to the second stage, and $\eta = 0$ signifies censoring after the first stage. The total survival time ($T^{unob}$) was defined as:

$$T^{unob} = T_1 + \eta * T_2,$$

where $T_2$ represents the second-stage outcome.

The observed total survival time $T = min(T^{unob}, C)$ is the minimum of the total survival time ($T^{unob}$) and the censoring time (C). The observed outcomes for each stage were defined as follows:

1. For the second stage, the observed outcome ($R_2$) was calculated as:

    $R_2 = (T - T_1) * I(T \geq T_1)$ if $\eta = 1$, else NaN

    Here if the patient does not enter stage 2, $R_2$ is set as undefined (NaN).

2. For the first stage, the observed outcome $R_1$ is defined as:

    $R_1 = T * I(T < T_1) + T_1 * I(T \geq T_1)$

To facilitate understanding of the notations used in results, we summarize key notations in Table 1. This framework for censoring and outcome generation ensures that the simulated data aligns with the complexities of survival analysis in multi-stage decision-making contexts. By integrating realistic censoring percentage and structured definitions of observed outcomes, this approach facilitates robust evaluation of the proposed method under various censoring scenarios. Note the learning algorithms will see only the observed information – here $T$, $R_1$, and $R_2$ – but not $T_1$ and $T_2$. As we have large enough samples, we split the data into 5 folds, using one fold for training and the other 4 for evaluation each time to ensure reliable performance estimation. This choice of using a single fold for training allows us to simulate a more challenging scenario, where the model is trained on limited data. Such an approach is particularly valuable in dynamic treatment regime optimization, as it helps evaluate the model's robustness and generalizability when applied to unseen data. Additionally, reserving a larger portion of the data for evaluation provides a more precise estimate of the model's performance by minimizing the variability in test results across folds.

Table 1: Key notations for semi-synthetic data generation process.

| Notation | Description |
|---|---|
| $A_1$ | Stage 1 treatment $A_1$ was assigned using a multinomial distribution. |
| $A_2$ | Stage 2 treatment $A_2$ was assigned using a multinomial distribution. |
| $g_1^{opt}$ | Stage 1 optimal treatment. |
| $g_2^{opt}$ | Stage 2 optimal treatment. |
| $T_1$ | Time spent in Stage 1 if no censoring happens. |
| $T_2$ | Time at Stage 2 if no censoring happens. |
| $T^{unob}$ | Total survival time if no censoring happens ($T_1 + T_2$). |
| C | Censoring time: The time at which a patient's survival is no longer observed. |
| T | Observed total survival time due to censoring: $min(T^{unob}, C)$. |
| $R_1$ | Survival time at Stage 1 after adjusting for censoring. |
| $R_2$ | Survival time at Stage 2 after adjusting for censoring. |

| η | $I(T_1 < C)$: Equals 1 if the patient enters Stage 2, otherwise 0. |

## Models for comparison

To evaluate the proposed CA-TRL method, we compared its performance against several approaches in both binary and multi-category treatment settings. Baseline models included fixed treatment policies, where treatments were uniformly assigned to all individuals (e.g., for g=0, assign treatment 0 to all patients; similarly for g=1, assign treatment 1, and so on as used in Table 2). A random treatment assignment policy, where treatments were allocated without considering individual covariates, was also included as a simplistic benchmark.

In addition to these baselines, we compared CA-TRL with the ASCL method [17], a recent advanced approach in dynamic treatment regime optimization. The ASCL method leverages a contrast-learning framework, which is based on the idea of the doubly-robust weighted classification approach. By incorporating pseudo-value approaches to address censoring in survival data, ASCL used penalized support vector machines. ASCL has demonstrated superior performance over earlier standard methods and is recognized for its adaptability and robustness in handling survival data. We selected ASCL as a benchmark because it is one of the most recent methods developed for censored dynamic treatment regimes and has a publicly available implementation, ensuring reproducibility and ease of comparison. We have used the publicly available implementation of ASCL.

The evaluation metrics included Restricted Mean Survival Time (RMST), which captures the average survival time up to a specified truncation point τ, making it well-suited for censored data. Correct Decision Rate at the First Stage (CDR1) measured the accuracy of treatment assignments at the initial stage, while Average Correct Decision Rate (ACDR) provided an overall assessment of decision accuracy across all stages [17]. Counterfactual outcomes were also analyzed to estimate the survival benefits that would have if individuals followed the treatments recommended by each method [11].

## Analysis Results

The evaluation results, presented in Table 2, demonstrate the superior performance of the proposed CA-TRL method across various evaluation metrics, including τ-restricted mean survival time (τ-RMST), CDR1, ACDR, and expected survival outcomes. These metrics were computed under different treatment options, propensity models, and censoring types.

For binary treatments, CA-TRL consistently outperformed all comparison methods, including random treatment assignment and the ASCL algorithm. Under exponential censoring with a true propensity model, CA-TRL achieved a τ-RMST of 46.37 ± 0.28 compared to 34.50 ± 3.15 for ASCL. Additionally, CA-TRL achieved a CDR1 of 93.17 ± 5.90 and an ACDR of 89.13 ± 2.40, both significantly higher than those obtained by ASCL, which recorded 70.80 ± 9.88 and 52.09 ± 7.36, respectively. Similar trends were observed for other censoring and propensity model types, with CA-TRL consistently demonstrating higher τ-RMST, CDR1, and ACDR values.

For three categorical treatment options, the results were equally compelling. Under conditional censoring with a false propensity model, CA-TRL achieved a τ-RMST of 26.11 ± 0.19, a CDR1 of 80.28 ± 4.25, and an ACDR of 77.28 ± 4.73. In contrast, ASCL showed significantly lower values with a τ-RMST of 14.35 ± 2.89, a CDR1 of 66.64 ± 13.6, and an ACDR of 20.09 ± 11.62. Similar trends were observed for other scenarios. These findings highlight the robustness and adaptability of the proposed method in complex multi-treatment scenarios.

CA-TRL's effectiveness is further validated by its ability to achieve higher survival outcomes across all scenarios. The expected survival outcomes under CA-TRL consistently outperformed those of other methods. For instance, under uniform censoring with a true propensity model for binary treatments, CA-TRL recorded an expected survival outcome of 92.45 ± 1.90 compared to 77.52 ± 1.77 for ASCL.

Overall, these results underline the strength of CA-TRL in accurately estimating dynamic treatment regimes and maximizing survival outcomes, regardless of treatment complexity, censoring type, or propensity model specifications. The consistent outperformance across various metrics establishes CA-TRL as a reliable method for personalized treatment planning.

Table 2: Performance of several DTR algorithms. The table reports the $\tau$-restricted survival time, correct decision rate at the first stage (CDR1), average correct decision across all stages (ACDR), and Expected Survival outcome. For each scenario, the best model is highlighted in bold.

| Treatment Options | Propensity Model Type | Censoring Type | Method | $\tau$ | $\tau$-RMST | CDR1 | ACDR | Expected Survival outcome $E[T^*\|g]$ |
|---|---|---|---|---|---|---|---|---|
| 2 | True | Exponential | g=0 | 52 | 16.16±0.12 | 80.14±0.06 | 66.19±0.04 | 24.25±0.05 |
| | | | g=1 | | 2.86±0.01 | 19.86±0.06 | 5.01±0.04 | 3.24±0.0 |
| | | | Random | | 9.34± 0.02 | 17.33±0.06 | 3.01±0.07 | 12.41±0.07 |

| | | | | | | | |
|---|---|---|---|---|---|---|---|
| | | | ASCL | | 34.50±3.15 | 70.80±9.88 | 52.09±7.36 | 62.12±8.43 |
| | | | CA-TRL | | **46.37±0.28** | **93.17±5.90** | **89.13±2.40** | **93.98±1.93** |
| | | Conditional | g=0 | 62 | 14.70±0.05 | 80.14±0.05 | 68.53±0.09 | 24.21±0.04 |
| | | | g=1 | | 2.75±0.01 | 19.86±0.05 | 3.91±0.02 | 3.24±0.00 |
| | | | Random | | 9.29±0.09 | 17.43±0.05 | 2.90±0.04 | 12.37±0.05 |
| | | | ASCL | | 40.80±3.06 | 68.11±9.38 | 54.60±6.67 | 57.25±4.24 |
| | | | CA-TRL | | **57.62±0.17** | **91.38±3.38** | **92.01±2.50** | **93.25±1.16** |
| | | Uniform | g=0 | 13 | 9.30±0.01 | 80.14±0.02 | 66.71±0.05 | 24.21±0.04 |
| | | | g=1 | | 2.82±0.0 | 19.86±0.02 | 4.80±0.2 | 3.24±0.0 |
| | | | Random | | 6.30±0.01 | 17.39±0.01 | 2.99±0.03 | 12.41±0.03 |
| | | | ASCL | | 12.62±0.01 | 86.03±1.44 | 78.72±3.10 | 77.52±1.77 |
| | | | CA-TRL | | **12.84±0.03** | **89.96±2.14** | **90.33±1.95** | **92.45±1.90** |
| | False | Exponential | g=0 | 52 | 16.19±0.09 | 80.14±0.03 | 66.22±0.05 | 24.21±0.05 |
| | | | g=1 | | 2.86±0.01 | 19.86±0.03 | 4.99±0.04 | 3.24±0.0 |
| | | | Random | | 9.32±0.03 | 17.42±0.08 | 3.08±0.06 | 12.37±0.05 |
| | | | ASCL | | 33.51±1.81 | 67.86±8.94 | 48.44±4.91 | 59.94±5.01 |
| | | | CA-TRL | | **46.87±0.35** | **95.77±3.60** | **93.34±2.89** | **95.79±1.53** |
| | | Conditional | g=0 | 62 | 14.67±0.04 | 80.16±0.05 | 68.48±0.09 | 24.20±0.05 |
| | | | g=1 | | 2.76±0.01 | 19.84±0.05 | 3.92±0.02 | 3.24±.01 |
| | | | Random | | 9.25±0.11 | 17.41±0.05 | 2.87±0.02 | 12.37±0.07 |
| | | | ASCL | | 41.23±3.75 | 71.33±10.47 | 54.57±7.42 | 58.63±5.11 |
| | | | CA-TRL | | **57.58±0.21** | **90.07±4.02** | **91.02±2.97** | **92.69±1.32** |
| | | Uniform | g=0 | 13 | 9.31±0.01 | 80.14±0.02 | 66.71±0.05 | 24.21±0.04 |
| | | | g=1 | | 2.82±0.0 | 19.86±0.02 | 4.80±0.02 | 3.24±.0.0 |
| | | | Random | | 6.30±0.01 | 17.40±0.01 | 2.99±0.03 | 12.41±0.03 |
| | | | ASCL | | 12.67±0.01 | 87.71±0.73 | 80.27±2.95 | 79.13±1.75 |
| | | | CA-TRL | | **12.84±0.03** | 89.87±2.20 | 90.25±1.78 | 92.42±1.90 |
| 3 | True | Exponential | g=0 | 23 | 7.98±0.04 | 80.14±0.03 | 62.01±0.09 | 8.45±0.01 |
| | | | g=1 | | 5.97±0.08 | 7.74±0.03 | 0.00±0.0 | 12.35±0.02 |
| | | | g=2 | | 1.38±0.0 | 12.12±0.04 | 4.78±0.03 | 1.60±0.0 |
| | | | Random | | 7.60±0.10 | 17.54±0.07 | 3.32±0.07 | 12.32±0.06 |
| | | | ASCL | | 13.29±0.97 | 75.12±4.76 | 21.17±11.90 | 28.91±8.54 |
| | | | CA-TRL | | **20.19±0.14** | 79.33±3.45 | **80.18±4.10** | **80.15±2.30** |
| | | Conditional | g=0 | 26 | 7.56±0.01 | 80.18±0.05 | 63.44±0.06 | 8.45±0.0 |
| | | | g=1 | | 4.56±0.02 | 7.72±0.03 | 0.0±0.0 | 12.36±0.06 |
| | | | g=2 | | 1.32±0.02 | 12.10±0.03 | 4.28±0.03 | 1.59±0.0 |
| | | | Random | | 7.69±0.11 | 17.48±0.08 | 3.34±0.08 | 12.36±0.05 |
| | | | ASCL | | 14.67±3.44 | 63.25±16.94 | 26.93±18.97 | 31.22±13.04 |
| | | | CA-TRL | | **24.82±0.42** | **82.77±2.70** | **83.03±2.36** | **80.95±1.70** |
| | | Uniform | g=0 | 5 | 4.15±0.0 | 80.11±0.03 | 62.35±0.07 | 8.45±0.01 |
| | | | g=1 | | 2.69±0.0 | 7.77±0.03 | 0.0±0.0 | 12.36±0.05 |
| | | | g=3 | | 0.95±0.0 | 12.12±0.02 | 4.80±0.02 | 1.60±0.0 |
| | | | Random | | 3.28±0.01 | 17.46±0.07 | 3.43±0.2 | 12.39±0.08 |
| | | | ASCL | | 4.28±0.0 | 80.11±0.03 | 0.0±0.0 | 12.44±0.01 |
| | | | CA-TRL | | **4.93±0.04** | **87.24±3.16** | **69.22±7.31** | **76.55±7.04** |
| | False | Exponential | g=0 | 23 | 8.02±0.03 | 80.13±0.04 | 62.01±0.11 | 8.45±0.0 |
| | | | g=1 | | 6.00±0.09 | 7.74±0.04 | 0.0±0.0 | 12.35±0.03 |
| | | | g=2 | | 1.38±0.01 | 12.13±0.04 | 4.78±0.03 | 1.60±0.0 |
| | | | Random | | 7.60±0.13 | 17.60±0.02 | 3.34±0.09 | 12.31±0.07 |
| | | | ASCL | | 13.5±1.23 | 74.20±5.93 | 23.91±15.02 | 30.42±10.89 |
| | | | CA-TRL | | **19.96±0.69** | **85.41±4.02** | **79.06±8.57** | **80.22±6.20** |
| | | Conditional | g=0 | 28 | 7.47±0.04 | 80.14±0.05 | 63.44±0.08 | 8.45±0.01 |

|   |   |   |   |   |   |   |   |
|---|---|---|---|---|---|---|---|
|   |   | g=1 |   | 4.44±0.04 | 7.74±0.04 | 0.0±0.0 | 12.35±0.02 |
|   |   | g=2 |   | 1.31±0.02 | 12.12±0.03 | 4.33±0.03 | 1.60±0.0 |
|   |   | Random |   | 7.67±0.06 | 17.48±0.07 | 3.22±0.05 | 12.41±0.04 |
|   |   | ASCL |   | 14.35±2.89 | 66.64±13.65 | 20.09±11.62 | 27.44±8.21 |
|   |   | CA-TRL |   | **26.11±0.19** | **80.28±4.25** | **77.28±4.73** | **79.01±2.51** |
|   | Uniform | g=0 | 5 | 4.15±0.0 | 80.14±0.04 | 62.40±0.07 | 8.45±0.01 |
|   |   | g=1 |   | 2.68±0.01 | 7.74±0.03 | 0.0±0.0 | 12.35±0.04 |
|   |   | g=2 |   | 0.95±0.0 | 12.12±0.02 | 4.8±0.01 | 1.60±0.0 |
|   |   | Random |   | 3.28±0.0 | 17.45±0.05 | 3.40±0.04 | 12.37±0.06 |
|   |   | ASCL |   | 4.33±0.05 | 80.88±0.77 | 6.69±6.69 | 17.43±4.99 |
|   |   | CA-TRL |   | **4.99±0.0** | **91.55±1.98** | **75.91±2.43** | **88.41±2.39** |

## Observational studies

The proposed method was applied to the SANAD dataset [33] to illustrate its utility in estimating DTRs in a real-world clinical setting. The SANAD study, a randomized controlled trial, evaluated the effectiveness of two antiepileptic drugs—carbamazepine (CBZ) and lamotrigine (LTG)—for patients with partial epilepsy. Treatment failure, defined as the withdrawal of the randomized drug due to unacceptable adverse effects (UAE) or inadequate seizure control (ISC), served as the primary outcome. The dataset includes both longitudinal and survival data for 605 patients, randomly assigned to CBZ ($A_1 = 1$) or LTG ($A_1 = 2$).

Our decision to use this dataset and the preprocessing steps were inspired by a study by Speth et al. work [15], which employed this data to explore DTR estimation. While the original SANAD study [33] is a randomized trial, the rate of drug titration—encompassing initial doses and subsequent dose adjustments—was determined by clinicians based on their judgment, effectively making it observational in nature. Speth et al. [15] highlighted that this setup provides a realistic scenario for evaluating DTRs, as the dosing adjustments introduce variability reflective of clinical practice.

Following Speth et al.'s approach, we focused on the first two visits, where clinicians could adjust the dose based on patient outcomes. At each visit ($k > 1$), the clinician could decrease ($A_k = 1$), maintain ($A_k = 2$), or increase ($A_k = 3$) the dose. Patients were followed up regularly at intervals determined by clinical judgment, leading to variability in follow-up schedules. Due to the limited number of observed events beyond the third visit, the analysis was restricted to the first two visits to ensure a reliable estimation of optimal treatment regimes. The preprocessing involved excluding 31 patients who did not receive treatment (zero initial and subsequent doses) from the analysis. This left 574 patients, with an event rate of approximately 33%.

Baseline covariates included age, gender, and the presence of a learning disability (recorded as "Yes" or "No"). Longitudinal data encompassed the calibrated dose of medication at each visit, along with the start and end times of treatment intervals. For stage 1 treatment prediction, we utilized age, gender, the start time of the dose interval, and the presence of a learning disability as input features. For stage 2 treatment prediction, we incorporated all stage 1 features along with the end time of the stage 1 interval and the corresponding dose amount.

Analysis Results

To evaluate the performance of the proposed model on this dataset, we employed 10-fold cross-validation due to the limited sample size. The proposed CA-TRL model demonstrated superior performance, achieving a restricted mean survival time (RMST) of 1742.67 with a standard error of 75.23, compared to ASCL, which achieved an RMST of 1509.12 with a standard error of 70.78. This difference is statistically significant ($p < 0.05$). Additionally, patients whose actual treatments aligned with the estimated optimal treatments recommended by CA-TRL exhibited significantly better survival outcomes than those whose treatments did not align with the recommendations ($p < 0.05$). At the initial visit, the estimated DTRs by CA-TRL suggest prescribing LTG to a greater number of individuals compared to the current treatment strategy. This observation aligns with prior findings [15,34], which highlight LTG's advantage over CBZ in reducing treatment failures, despite CBZ being the long-standing standard treatment.

# Discussion and Future Scope

This work presents CA-TRL, a censoring-aware extension of tree-based reinforcement learning (T-RL), which addresses the critical need for robust estimation of dynamic treatment regimes (DTRs) in survival analysis. By integrating censoring adjustments and leveraging augmented inverse probability weighting (AIPW), CA-TRL successfully navigates challenges associated with incomplete data. The model's performance was validated through rigorous simulations and real-world application, particularly on the SANAD dataset, where it outperformed state-of-the-art methods such as ASCL in key metrics like $\tau$-restricted mean survival time ($\tau$-RMST) and decision accuracy. These results affirm CA-TRL's potential to guide personalized treatment strategies in clinical practice.

Despite its strengths, CA-TRL has certain limitations that warrant further research. First, the framework currently supports discrete treatment options, limiting its applicability to settings where treatments can be categorized into distinct groups. However, many real-world scenarios, such as drug dosage optimization, involve continuous treatment variables. Second, treatment selection bias remains a challenge, particularly in observational datasets where treatment assignments may reflect underlying clinician preferences. While CA-TRL incorporates propensity score adjustments to mitigate confounding, residual bias could influence the estimated treatment regimes. Incorporating more sophisticated causal inference techniques could further strengthen the model's robustness and reliability. Third, it employs a greedy approach when selecting splits, focusing solely on the purity improvement at the current node. While this strategy simplifies the tree construction process, it may not always lead to the global optimum in terms of overall decision rule performance. An alternative approach to enhance the performance of this method is to incorporate a "lookahead" mechanism [35]. By evaluating potential future splits and their impact on downstream nodes, this method can identify splits that optimize the tree structure more holistically, potentially leading to better overall treatment decisions. Finally, while the method demonstrated promising results on the SANAD dataset, additional validation across diverse clinical conditions is necessary to generalize its applicability. Future studies could explore CA-TRL's performance in domains such as cardiovascular diseases, and diabetes management, refining the methodology for specific clinical settings.

# Conclusions

CA-TRL offers a robust and interpretable solution for estimating dynamic treatment regimes in settings with censored survival outcomes. By integrating censoring-aware adjustments and leveraging the strengths of tree-based methodologies, it provides an effective framework for personalized medicine. The model's superior performance in both simulated and real-world datasets highlights its potential to improve clinical decision-making. However, addressing limitations related to continuous treatments, and treatment selection bias will be critical for broader adoption. By pursuing these advancements and validating the approach across diverse medical domains, CA-TRL can serve as a transformative tool in precision medicine, fostering improved patient outcomes and personalized care strategies.

# Methods

## a. Notation and assumptions

We use the context of coronary artery disease (CAD) as a motivating example to illustrate the dynamic treatment regimes (DTR) framework. CAD is a leading cause of hospitalization and mortality worldwide, presenting a complex clinical challenge that necessitates personalized and adaptive treatment strategies to optimize survival outcomes. We focus on CAD in this study due to its clinical significance, high prevalence, and the availability of detailed patient data in the MIMIC-IV dataset [18]. In our study, we assume treatment decisions are made over two stages to maximize the long-term outcome (the patient's overall survival time): Stage 1, focusing on renal and anemia management, followed by stage 2, focusing on glycemic management. These stages reflect two critical aspects of CAD management that affect long-term outcomes. The event of interest is defined as a critical clinical endpoint (e.g., mortality) that could terminate follow-up. It is important to note that this example is designed solely for experimental purposes, and actual treatment stages and options may differ in real-life clinical scenarios.

We assume that data from $N$ individuals are available. The data from the individuals are independent, identically distributed (iid) copies of covariates ($X_k$), treatments ($A_k$), and outcomes ($R_k$) for all individuals, at each stage $k$. In the CAD context, X includes patient clinical measurements such as age, creatinine, hemoglobin, potassium, sodium, glucose, platelet count, hematocrit, and white blood cell count. The treatment indicator $A_k$ can represent a multi-categorical or ordinal treatment, with observed values $a_k \in A_k^{options}$, where $|A_k^{options}| = M_k \geq 2$ denotes the number of available treatment options at the stage $k$. For instance, in the CAD context, stage 1 focuses on renal and anemia management, with treatment options including conservative monitoring (observation without intervention), moderate support (medications or dietary adjustments for mild dysfunction), and aggressive intervention (treatment for severe anemia or renal dysfunction). If this stage finishes, the patient moves to Stage 2, which addresses glycemic management, with several treatment options: conservative glycemic management (dietary and lifestyle interventions), pharmacological intervention (glucose-lowering medications), versus aggressive glycemic control (intensive management using insulin or combination therapies). In this framework, $R$ represents the overall observed survival time, defined as $R = \sum_{k=1}^{K} R_k$, where $R_k$ denotes the observed survival time at the stage $k$ until the next stage or the event of interest. This serves as a quantitative measure for evaluating the effectiveness of treatment policies across the two stages.

Individual data are represented as sequences of covariates, treatments, and outcomes [ [$X_1$, $A_1$, $R_1$], ..., [$X_K$, $A_K$, $R_K$] ], organized by stages, with each stage indicated by the subscript $k$. Here, stages refer to distinct points in the treatment process where decisions are made about interventions. Let individuals be identified with a superscript $i \in \{1, ..., N\}$, although this subscript is often omitted for simplicity. For example, instead of explicitly writing [$X_k^i$, $A_k^i$, $R_k^i$] for the covariates, treatments, and outcomes of the individual $i$ at the stage $k$, we often simplify notation by using [$X_k$, $A_k$, $R_k$], implicitly assuming that the data are indexed by $i$ when referring to a single individual.

The policy can use the values of the covariates $X_k$ as they are known at the start of stage $k$, to decide which treatment $A_k$ to administer. This means the treatment options for the second stage can be influenced by the treatment given in the first stage, as those second-stage covariates are determined by the treatment $A_1$, covariates $X_1$, and outcome observed during the first stage $R_1$.

Let $h_k$ represent the histories, which include (a function of) covariates, previous treatments, and previous outcomes available at the start of the stage $k$, used to guide the treatment decision for that stage. For instance, $h_k$ could be $h_k = f_k(X_1, A_1, R_1, ..., X_{k-1}, A_{k-1}, R_{k-1}, X_k)$, where $f_k(.)$ is a function summarizing past information at stage $k$. For example, perhaps $h_1 = X_1$ for the first stage and $h_2 = (X_1, A_1, R_1, X_2)$ for the second stage, etc. In this paper, we specifically assume $f_k(.)$ to be the identity function, such that $f_k(Q) = Q$ for all $k$. This means that the history $h_k$ directly incorporates all covariates, treatments, and outcomes from prior stages up to stage $k$, without any transformation or aggregation.

In the CAD context, at stage 1, the patient's history $h_1$ includes baseline covariates such as age, and laboratory values like creatinine and hemoglobin levels. Based on this history, the policy makes a treatment decision $A_1$, such as prescribing a specific renal management strategy. The observed survival time during stage 1 is denoted by $T_1 \in \Re$ representing the time until the next decision point, which could be either another treatment stage or terminating the process (perhaps due to an event like kidney failure, or death). If there is no terminal event during stage 1, the policy then always advances to stage 2. Here, the details of that next decision point are influenced by the patient's clinical progression and the effectiveness of the current treatment. For example, this waiting period for CAD patients could be clinically determined based on the patient's response to the initial treatment, the stability of their renal function, and other indicators such as hemoglobin levels. At this point, the policy uses the updated history $h_2$, which incorporates the treatment from stage 1 $A_1$, the elapsed time $T_1$ during stage 1, and additional covariates $X_2$ such as glucose levels at the end of stage 1 (before the start of stage 2). It can then use this $h_2$ to decide on the next treatment action – e.g., to help guide decisions for glycemic management. Finally, we can have the observed stage-2 survival time $T_2 \in \Re$, representing the time until terminating the process (perhaps due to an event like kidney failure). The overall survival time for an individual is then defined as $T = T_1 + T_2$.

However, not all patients experience the event of interest within the observation period $\tau > 0$, resulting in censored outcomes. Censoring can occur, for example, when the study concludes before a patient experiences the event (e.g., when $T > \tau$) or when a patient is lost to follow-up, etc. When $T > \tau$, no information on survival is available beyond the study length $\tau$ in the observed data, meaning the outcome of interest is truncated by $\tau$. The overall censoring time $C$ marks the point when the observation of an individual's survival time stops without the event of interest occurring. If the censoring happens during an earlier stage, the patient does not progress to the next stage, which is another way the survival time $T$ may be censored for some CAD patients.

Formally, let $\eta_k$ be a random variable that equals 1 if an individual is known to enter stage k and 0 otherwise, with the condition that $\eta_1 = 1$ for all individuals (as we assume that no individual can be censored before entering stage 1). Additionally, If $\eta_k = 0$, then for any future stage $j > k$, $\eta_j = 0$. The duration of survival during each stage $k$ is represented by $T_k$, defined as the time interval from the time the individual entered stage k until they either progress to stage $k + 1$ or experience a terminating event such as failure or death, whichever occurs first. For an individual, this process generates a sequence of covariates, treatments, and outcomes [ [ $X_1, A_1, R_1$], ..., [$X_K, A_K, R_K$] ] ], where the observed survival outcome is $R_k = min(T_k, C_k)$, and $C_k$ represents the remaining censoring time at the stage $k$, calculated based on the individual's overall censoring time $C$ and their progression through the stages. Specifically, $C_k$ is defined as $C_k = C - \sum_{j=1}^{k-1} \eta_{j+1} T_j$, where $\sum_{j=1}^{k-1} \eta_{j+1} T_j$ represents the cumulative survival time up to stage $k - 1$. The censoring indicator $\delta_k = I(T_k \leq C_k)$ specifies whether the event occurred without censoring before the end of the stage $k$. If censoring occurs at a given stage, no further outcomes, treatments, or covariates are observed for subsequent stages – ie, this process just terminates. The observed outcome is defined as the overall survival time, which is the sum of the survival times across all stages, $T = \sum_{k=1}^{K} \eta_k R_k$.

A Dynamic Treatment Regime (DTR) consists of a series of decision rules $g = [g_1(h_1),..., g_K(h_K)] \in G$, where G represents the set of all possible treatment regimes. Each policy $g$ comprises stage-specific decision rule functions $[g_1, g_2,..., g_K]$. At each stage $k$, the decision rule $g_k(.) \in G_k$ is a function that maps the history $h_k$ to a treatment, where $G_k$ represents the set of all possible decision rules at the stage $k$. This means that $g_k(h_k)$ provides a treatment recommendation based on the history up to stage $k$.

The goal of an optimal DTR is to maximize the expected patient outcomes over the entire treatment course. To define this mathematically, we adopt a counterfactual framework. While $T$ captures the observed survival through all K stages, $T^*$ represents the potential (counterfactual) survival outcome under another tuple of hypothetical treatment assignments. At the final stage K, let $T^*(a_1,..., a_K)$ represent the counterfactual outcome for a patient who received the treatments $a_1,..., a_K$. For a specific treatment rule $g_K$, the counterfactual outcome $T^*(g_K)$ is given by:

$$T^*(g_K) = \sum_{a_K=1}^{M_K} T^*(a_k) \times I\{g_K(H_K) == a_K\}$$

where $I\{g_K(H_K) == a_K\}$ is an indicator function that equals 1 if the treatment prescribed by $g_K(H_K)$ matches $a_K$. The performance of a treatment regime $g$ is measured by the value function, $V(g) = E[T^*(g)]$, which represents the expected counterfactual outcome for all individuals if they followed the regime $g$. The optimal regime, denoted by $g^{opt}$, is the one that maximizes this value function: $V(g^{opt}) \geq V(g), \forall g \in G$. For example, in the case of the aforementioned CAD example, the value function, $V(g)$, represents the expected survival time if all patients follow a specific regime $g$. The optimal regime, $g^{opt}$, maximizes $V(g)$, ensuring the longest expected survival across the population.

Since only one of the counterfactual outcomes is observed for each individual, estimating the full counterfactual distribution requires specific assumptions. To address this challenge, we rely on the below key assumptions [11,15]:

- Positivity: This condition, $P(A_k = a_k | H_k = h_k) > 0, \forall a_k \in A_k, \forall h_k)$, ensures that all individuals have a chance of receiving each treatment. In practice, violations commonly arise in the empirical sample when certain treatments are never (or almost never) prescribed to individuals with specific characteristics (e.g., sicker patients rarely receive a more intensive therapy). This can lead to regions of the covariate space where one treatment is effectively unavailable, making estimation of counterfactuals infeasible.
- Consistency: The observed outcome for an individual under a given treatment matches the counterfactual outcome predicted under that treatment.
- No unmeasured confounding: It ensures that, given the observed covariates, the treatment assignment is independent of the potential outcomes, or stated another way, the outcome of a treatment depends on the observed covariates, but NOT on the decision of which treatment to apply. For instance, for stage 1, $R_1^*(a_1) \perp A_1 | H_1$, for stage 2, $R_2^*(a_1, a_2) \perp A_2 | H_2$, and so on, where $R_1^*(a_1)$ and $R_2^*(a_1, a_2)$ represent the potential outcomes corresponding to the respective treatment actions. This assumption enables valid causal inferences and is sometimes referred to as 'strongly ignorable treatment assignment' or 'exchangeability'. For the longitudinal setting, this assumption requires that the treatment assignment at a given stage cannot depend on future covariates and treatment history. It is also called 'sequential randomization'. This assumption is frequently the biggest challenge in observational data, where there may be variables related to both treatment decisions and outcomes that are not measured (unmeasured confounders). Violations can lead to biased estimates of causal effects.
- No inference assumption: The outcome for any individual should not be influenced by the treatment assignment of another individual. It is also called 'Stable unit treatment value'. It can be difficult to maintain in real-world clinical settings where doctors' decisions for one patient often influence subsequent treatments and outcomes.

By adhering to these assumptions, the observed data can be linked to the counterfactual framework, allowing for valid causal inference and identification of optimal DTRs.

Under these assumptions, the optimal decision rule $g_k^{opt}$ at any stage k is derived recursively, starting from the final stage and moving backward. This is defined as:

Final Stage $K$: The optimal rule $g_K^{opt}$ at the last stage K is

$$g_K^{opt} = \arg\max_{g_K \in G_K} E[\sum_{a_K=1}^{M_T} E(T|A_K = a_K, H_K)I(g_K(H_K) == a_K)]$$

Intermediate Stages ($k = 1, \ldots, K - 1$): For earlier stages, the optimal rule $g_k^{opt}$ is defined recursively, incorporating the expected outcomes of future stages. We consider $T^*(A_1, \ldots, A_{k-1}, g_k, g_{k+1}^{opt}, \ldots, g_K^{opt})$, which represents the counterfactual outcome if a patient follows the decision rule $g_k$ at stage k, assuming that all future stages $(k+1, \ldots, K)$ are governed by the optimal rules $g_{k+1}^{opt}, \ldots, g_K^{opt}$, given the treatments $A_1, \ldots, A_{k-1}$ up to stage $k - 1$.

At any intermediate stage $k$, the optimal decision rule is given by:

$$g_k^{opt} = \arg\max_{g_k \in G_k} E[T^*(A_1, \ldots, A_{k-1}, g_k, g_{k+1}^{opt}, \ldots, g_K^{opt})]$$

$$= \arg\max_{g_k \in G_k} E[\sum_{a_k=1}^{M_k} E[\overline{R}_k|A_{k+1} = a_k, H_k)I\{g_K(H_k) == a_k\}]]$$

where, $\overline{R}_k$ is the pseudo outcome at stage $k$, representing an intermediate estimate of the counterfactual outcome based on observed data and accounting for the optimal decisions at future stages.

## b. Censoring-Aware Tree-Based Reinforcement Learning (CA-TRL)

Censoring-aware tree-based reinforcement learning builds on the foundational principles of decision trees and is inspired by the work of Tao et al. [11], who extended traditional tree-based methods to optimize counterfactual outcomes in dynamic treatment regimes (DTRs). Tao et al.'s framework focuses on complete datasets without censoring, where all outcomes are fully observed. In contrast, CA-TRL adapts this approach to address the complexities of right-censored survival data, making it applicable in settings where incomplete observations are prevalent.

## Tree-Based Learning Framework

The construction of decision trees in CA-TRL follows the idea of recursively partitioning the covariate space to create regions that maximize homogeneity based on a "purity" criterion. Unlike standard decision tree methods, such as Classification and Regression Trees (CART) [36], which rely on directly observed outcomes, CA-TRL focuses on counterfactual mean outcomes. These represent the expected results under hypothetical treatment assignments and serve as the foundation for assessing purity.

At each stage of the tree-growing process, CA-TRL evaluates possible splits by calculating the counterfactual mean outcomes for the resulting child nodes. The purity improvement from a split is determined by comparing the counterfactual outcomes of the child nodes to the parent node. The split that maximizes this improvement is selected, ensuring that the tree partitions the covariate space in a way that optimizes treatment assignment.

**Recursive Partitioning for Counterfactual Optimization**

To guide the tree-growing process, CA-TRL employs a recursive partitioning framework. Let $\phi$ denote a parent node and $\omega$ a potential split at that node. The counterfactual purity measure evaluates the counterfactual mean outcome for the split and without splitting. The optimal split $\omega^*$ is selected to maximize the purity improvement. This recursive process continues until predefined stopping criteria are met, resulting in a tree structure that optimally partitions the covariate space to guide treatment decisions.

**Counterfactual Framework and Recursive Optimization**

The recursive optimization framework of CA-TRL ensures that decisions at each stage account for their impact on future outcomes. At stage $k$, the counterfactual outcome is expressed as:

$$T^*(A_1, \cdots, A_{k-1}, g_k, g^{opt}_{k+1}, \cdots, g^{opt}_T)$$

representing the expected outcome for an individual following the decision rule $g_k$ at stage $k$, assuming optimal decision rules are applied at all subsequent stages. This recursive structure aligns with Bellman's principle of optimality and provides a framework for dynamically refining treatment rules based on evolving patient histories.

CA-TRL incorporates pseudo-outcomes $\overline{R}_k$ at each stage to estimate counterfactual outcomes, particularly in the presence of censoring. These pseudo-outcomes are adjusted to account for incomplete data and serve as substitutes for the unobserved counterfactual outcomes. By integrating these censoring-aware adjustments, as described in the later section on CA-TRL estimation for stage $k$, CA-TRL ensures accurate and robust optimization of treatment rules.

**Stopping Rules to Prevent Overfitting**

CA-TRL employs stopping rules to prevent overfitting, following the framework of Tao et al.[11], These rules include:

1. **Minimum Node Size:** Nodes with fewer than $2n_0$ number of observations are not split, where $n_0$ denotes the minimum node size.
2. **Purity Improvement Threshold:** Splits resulting in a minimal improvement in purity (below a threshold λ) are disallowed.
3. **Maximum Tree Depth:** The tree-growing process is capped at a user-defined maximum depth.
4. **Child Node Size:** Splits that create child nodes smaller than $n_0$ observations are not performed.

If none of these stopping rules are triggered, the node is split, and the tree continues to grow until the criteria are satisfied.

## CA-TRL estimation for the final stage $K$

To address the causal objective of optimal DTR estimation at the final stage $K$, we propose a modified augmented inverse probability weighting estimator, specifically adapted for censored data. This approach builds upon the work of Tao et al. [11] and extends it to handle incomplete observations, ensuring robustness and accuracy in estimating counterfactual outcomes. The estimator for the counterfactual mean outcome at the final stage is given by

$$\hat{\mu}^{CAIPW}_{K,a_K}(H_K) = \frac{I(A==a_K)\delta_K}{\hat{\pi}_{a_K}(H_K)\hat{S}_C(T|H_K)}T + \left(1 - \frac{I(A==a_K)}{\hat{\pi}_{a_K}(H_K)}\right)\hat{\mu}_{K,a_K}(H_K)$$

where:

- $\delta_K$ is the censoring indicator,
- $\hat{\pi}_{a_K}(H_K)$ is the estimated propensity score for treatment $a_K$,
- $S_C(T|H_K)$ is the censoring survival probability conditional on the history $H_K$,
- $\hat{\mu}_{K,a_K}(H_K)$ represents the conditional mean outcome,
- $T$ is the observed outcome (the overall survival time across all stages).

This estimator exhibits double robustness, meaning it remains consistent if either the propensity model $\hat{\pi}_{a_k}(H_K)$ or the conditional mean model $\hat{\mu}_{K,a_K}(H_K)$ is correctly specified. The censoring survival probability $S_C(T|H_K)$ in the first term adjusts for the incomplete nature of the observed outcomes T, ensuring unbiased estimation. The second term does not require this adjustment, as the conditional mean model, such as random survival forests, directly accounts for censoring within its estimation.

## CA-TRL estimation for stages $1,..., K-1$

For intermediate stages ($K - 1 \geq k \geq 1$), the estimation of dynamic treatment regimes involves a recursive process that propagates information from future stages back to earlier stages. This process relies on pseudo-outcomes that capture the cumulative effects of treatment decisions.

We employ an adjusted version of pseudo-outcomes to mitigate the cumulative bias arising from the conditional mean models, as described by Huang et al. [37]. For complete data, rather than relying solely on model-derived values under optimal future treatments ($\hat{\mu}_{k+1, g^{opt}_{k+1}}(H_{k+1})$), we incorporate observed outcomes along with the anticipated loss resulting from suboptimal treatments. In the case of censored data, the estimation is based exclusively on $\hat{\mu}_{k+1, g^{opt}_{k+1}}(H_{k+1})$.

The pseudo-outcome $\overline{R}_k$ at stage $k$ is defined as:

$$\overline{R}_k = \delta_{k+1}(\overline{R}_{k+1} + \hat{\mu}_{k+1, g^{opt}_{k+1}}(H_{k+1}) - \hat{\mu}_{k+1, A_{k+1}}(H_{k+1})) + (1 - \delta_{k+1})\hat{\mu}_{k+1, g^{opt}_{k+1}}(H_{k+1})$$

where:

- $\delta_{k+1}$: Censoring indicator for stage $k + 1$ ($\delta_{k+1}$=1 if uncensored, 0 otherwise),
- $\overline{R}_k$: Pseudo-outcome at the stage $k + 1$,
- $\hat{\mu}_{k+1, g^{opt}_{k+1}}(H_{k+1})$ : Expected outcome under the optimal treatment rule at the stage $k + 1$,
- $\hat{\mu}_{k+1, A_{k+1}}(H_{k+1})$: Expected outcome under the observed treatment $A_{k+1}$ at the stage $k + 1$.

The estimator for the counterfactual mean outcome at the $k$ stage is given by

$$\hat{\mu}^{CAIPW}_{k, a_k}(H_k) = \frac{I(A==a_k)\delta_k}{\hat{\pi}_{a_k}(H_k)\hat{S}_C(\overline{R}_k|H_k)}\overline{R}_k + \left(1 - \frac{I(A==a_k)}{\hat{\pi}_{a_k}(H_k)}\right)\hat{\mu}_{k, a_k}(H_k)$$

## Implementation of CA-TRL

The implementation of CA-TRL begins with the stage $K$, where the observed outcome $T$ is directly used as the pseudo-outcome ($\bar{R}_K = T$), and proceeds via backward induction through earlier stages. The propensity scores ($\hat{\pi}_{a_k}(H_k)$) are estimated using multinomial logistic regression, while the censoring survival probability $S_C(T|H_K)$ and the conditional mean outcome ($\hat{\mu}_{K,a_k}(H_k)$) are estimated using random survival forests. The pseudo-outcomes ($\bar{R}_k$) are updated recursively, incorporating adjustments for censoring and the expected loss from suboptimal treatments, ensuring the robustness of the recursive estimation process and enabling effective handling of incomplete data. To optimize the performance of the proposed CA-TRL method, grid search was employed for hyperparameter tuning. The hyperparameters—number of trees, maximum tree depth, regularization parameter ($\lambda$), and minimum split size—were systematically varied. The optimal set of hyperparameters was selected based on the highest restricted mean survival time (RMST) value. This approach ensured robust and efficient tree-based estimation tailored to the dataset characteristics.